\PassOptionsToPackage{table,xcdraw}{xcolor}
\documentclass[sigconf]{acmart}
\usepackage{graphicx}
\usepackage[table,xcdraw]{xcolor}

\usepackage{pgfplots}
\usepackage{algorithmic}
\usepackage{algorithm}
\usepackage{amsmath}

\usepackage{amsfonts}
\usepackage{balance}
 \usepackage{booktabs}  
\usepackage{epsfig}
\usepackage{multirow}
\usepackage{xspace}
\usepackage{url}
\usepackage{color}
\usepackage {balance}
\usepackage{eqparbox}
\usepackage{mathrsfs}
\usepackage{caption}
\usepackage{subcaption}
\usepackage{lipsum}
\usepackage{enumitem}
\usepackage{float}
\usepackage{array, multirow}
\usepackage{balance}
\usepackage{hyperref}
\usepackage{mathtools}
\newcommand{\defeq}{\triangleq}

\usepackage{tabularx}
\usepackage{makecell}

\usepackage{soul}

\newcommand{\feicomment}[1]{{\color{orange} }}
\newcolumntype{L}[1]{>{\raggedright\let\newline\\\arraybackslash\hspace{0pt}}m{#1}}
\newcolumntype{C}[1]{>{\centering\let\newline\\\arraybackslash\hspace{0pt}}m{#1}}
\newcolumntype{R}[1]{>{\raggedleft\let\newline\\\arraybackslash\hspace{0pt}}m{#1}}

\newtheorem{Theorem}{Theorem}
\newtheorem{Definition}{Definition}

\newcommand{\eat}[1]{}

\definecolor{L_color}{rgb}{0.8,0.2,0.2}
\definecolor{Y_color}{rgb}{1.0, 0.0, 0.5}

\newcommand{\nop}[1]{}
\newcolumntype{Z}{ >{\centering\arraybackslash}X }

\newcommand{\yifan}[1]{}
\newcommand{\yifann}[2]{}

\begin{document}
\title{
SuperCone: 
Unified User Segmentation over Heterogeneous Experts via Concept Meta-learning
\\
}

\author{
Keqian Li$^{\dagger}$, Yifan Hu$^{\dagger}$\\
Yahoo Research$^{\dagger}$
}

\newcommand{\shortauthor}{K. Li et al.}

\renewcommand{\shorttitle}{SuperCone}

\begin{abstract}

We study the problem of user segmentation: given a set of users and one or more predefined groups or segments, assign users to their corresponding segments.
As an example, for a segment indicating particular interest in a certain area of sports or entertainment,
the task will be to predict whether each single user will belong to the segment.
However, there may exist numerous long tail prediction tasks that suffer from data availability and may be of heterogeneous nature, 
which make it hard to capture using single off the shelf model architectures.
In this work, we present \textit{SuperCone}, our unified predicative segments system that addresses the above challenges.
It builds on top  of a flat concept representation \cite{li2021hadoop, li2022metacon} that summarizes each user's heterogeneous digital footprints, 
and uniformly models each of the prediction task using an approach called "super learning ",
that is, combining prediction models with  diverse architectures or learning method that are not compatible with each other.
Following this, we provide an end to end approach that learns to flexibly attend to best suited heterogeneous experts adaptively,
while at the same time incorporating deep representations of the input concepts that augments the above experts.
Experiments show that \textit{SuperCone} significantly outperform state-of-the-art recommendation and ranking algorithms on a wide range of predicative segment tasks and public structured data learning benchmarks.\end{abstract}

\maketitle

\newcommand{\madelon}{\texttt{madelon}}
\newcommand{\ana}{\texttt{a9a}}

\begin{figure}[]
\centering
\includegraphics[width=.5\textwidth]{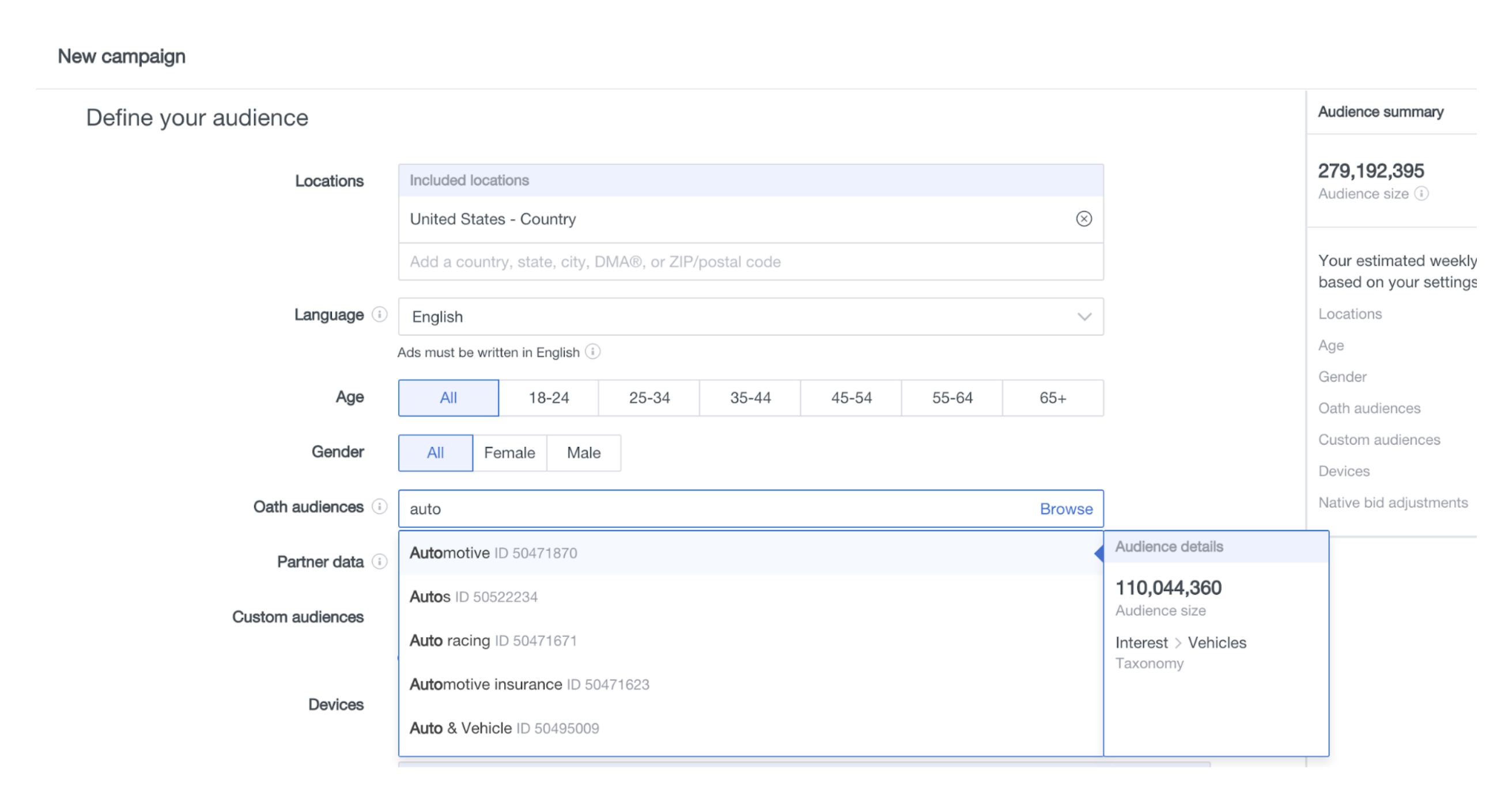}
\vspace{-10pt}
\caption{Illustration of \textit{SuperCone} use cases.
Fine-grained segments are provided to users for site customization and to
online advertisers for delivering effective content.
}
\label{fig-production system interface}
\vspace{-15pt}
\end{figure}

\section{Introduction}
We study the following problem: 
\textit{given a set of users and one or more predefined groups or segments, assign users to their corresponding segments based on the users' available online activities and existing membership between users and segments.} As a long-standing problem that has been in online information exchange with significant social and economic impact \cite{li2022}, it has played a pivotal role in supporting the continued growth of the platform that allow users to customize their content and for
the online advertisers to effectively budget their campaign.
\cite{cahill1997target}
\footnote{\url{https://www.facebook.com/business/ads/ad-targeting}} \footnote{\url{https://support.google.com/google-ads/answer/2497941?hl=en}}.

Typical user segmentation systems are constructed as follows. As illustrated in \autoref{fig-production system interface}, consumer and advertiser may log on to the system and indicate their preference of content, either in an explicit or implicit fashion. Large scale ranking and recommendation model \cite{zhao2019recommending, tang2020progressive} may follow that are designed for particular prediction tasks. However, existing approaches are not suitable for applying directly for the vast amount of possible prediction  tasks because of the reasons:
\begin{enumerate}
    \item  \textbf{Task Heterogeneity}:
    Digital footprints such as users' online activities may be logged and integrated  from a wide range of contexts and physical machine types
with large variations in terms of modality and schema, making it hard for learning system to adapt.
    \item  \textbf{Long-tailness}: The fine granularity of  segments that benefit information customization also results in
extremely large number of prediction tasks,
many of which belongs to the 
 long tail of the distribution with
 insufficient signal or observations.
    \item  \textbf{Data Availability}:
A large majority of the  of available learning signals are \textit{implicit feedback} \cite{hu2008collaborative} in nature and may suffer
\textit{missing at random} (MAR) effect.
In addition, the increasing awareness of user privacy
and compliance to regulation such as GDPR and platform constraints
such as \textit{Chromageddon}  \cite{chromium},
 \cite{voigt2017eu}
 \cite{iqbal2021towards}
 \textit{App Tracking Transparency} \footnote{\url{https://developer.apple.com/documentation/apptrackingtransparency}} and \textit{Intelligent Tracking Prevention}
all contribute to the availability of data.
Algorithm that relies on rich and complete features will suffer from the decrease in efficiency or effectiveness.
\end{enumerate}

\begin{figure*}[]
\centering
\includegraphics[width=\textwidth]{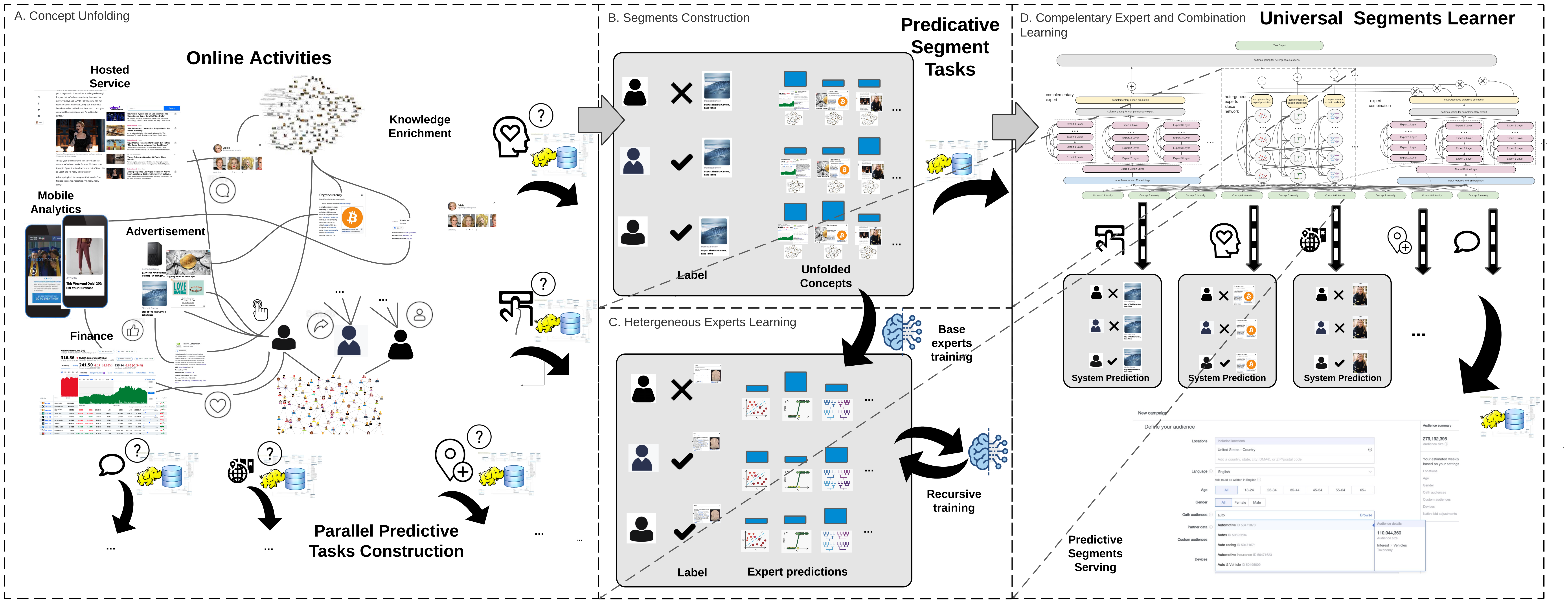}
\caption{ Overview of \textit{SuperCone} unified user segmentation system.
Online activities are integrated from heterogeneous sources where unfolded representation for each prediction tasks are constructed.
 A unified learning paradigm is then applied across different types of tasks by training heterogeneous experts and learning to combine them adaptively.
\yifan{This figure is too busy for the user to understand. Consider simplify? What is ETL?? Online serving-$>$ online activities? delete ``online behaviors" Explain what happen in each stage? E.g., task construction?}}
\label{fig-illustration}
\vspace{-15pt}
\end{figure*}
To address the above challenges,
we present \textit{SuperCone},
our unified approach for user segmentation that is able to apply to all prediction tasks in a consistent and reliable manner.
It builds the distributed concept representation \cite{li2021automl, lake2015human, li2022metacon} in order to obtain reliable representation of signal from   heterogeneous signals for each prediction task,
and model each of the tasks 
by combining heterogeneous prediction models that varies in  architectures or even learning method following the super learning paradigm \cite{naimi2018stacked}, 
while at the same time flexibly incorporating adaptive expert combination module
  and deep representations learning module from original input to augmenting the heterogeneous experts.
  We then provide an end-to-end approach
  for jointly learning the heterogeneous experts,
  the expert combination module, and the representations learning module under a principled meta learning framework.

Our contribution can be summarized as follows.
\begin{itemize}
     \item   We propose a unified solution to the critical problem of user segmentation,  \textit{SuperCone}, as an end-to-end solution that efficiently learns and combines arbitrary heterogeneous prediction models that are trained under a principled meta learning framework with provable performance advantages over other possible learning systems.
     \item   We conduct extensive evaluations of  \textit{SuperCone} over numerous user  segmentation task to demonstrate its substantial performance gain over the state-of-the-art recommendation and ranking approaches, the previous production system.
     \item  We apply \textit{SuperCone} to structured data learning problem and reports it superior performance on several public benchmark data-sets to further demonstrates the generalization of our approach.

\end{itemize}


\section{Related Work}
In this section,
we discuss key related work in relations to \textit{SuperCone} for the four following categories:
industrial ranking and recommendation system,
concept learning,
, meta learning
and super learning.

\subsection{Industrial Ranking and Recommendation System}
Industrial ranking and recommendation system are key to many aspects of internet business in areas such as  including advertising \cite{mcmahan2013ad}
and content serving \cite{covington2016deep},
with the majority of the approaches following the point wise learning-to-rank paradigm \cite{covington2016deep}.
For example, a ranking model may serve the resulted recommendation according to the predicted likelihood of engagement
\cite{tang2020progressive} such as satisfaction or end conversion.
Our work differs from the above by following a novel super learning architecture to incorporate heterogeneous experts
and apply it in the user segmentation scenario with low data availability and long tail tasks.

\subsection{Concept Learning}
The research on concept learning \cite{lake2015human}
focuses on mining concepts
from heterogeneous sources such as
relational data and semi-structured data lake \cite{wang2015concept} or
from unstructured data such as text documents \cite{li2019mining}.
Instead of unsupervised mining,  \cite{li2018unsupervised, li2014social, li2018concept, li2019hiercon, li2017discovering},
more recent work focuses specifically on obtaining reliable flat concept representation
from heterogeneous information such as user's implicit feedback \cite{li2021automl}
for large scale distributed learning scenarios.
Our approach builds on top of previous work and
further learns the interplay across downstream learners following the super learning framework.


\subsection{Ensemble Learning}
The research on ensemble learning \cite{dong2020survey} focuses on leveraging multiple machine learning models, commonly referred to as "experts".
A particular branch known as "super learning" in statistics \cite{naimi2018stacked}
focuses on training machine learning model such as a regression model for combining the predictions of individual "experts" into the final prediction.
Our method bring a couple of theoretical and practical advancement in this area,
including a much more generalized scheme for combining the individual models with provable optimality guarantee and a novel architecture for deep representation learning over "experts".

\subsection{Meta Learning}
The research of meta learning, also known as "learning to learning",  focuses on
the learning mechanism that gains experience and improves its performance over multiple learning episodes \cite{thrun1998learning}.
One of the most general class of all is architectural search,
where multiple instantiation of the model are learned jointly, with most performing ones being kept
and unfitted ones being discarded \cite{liu2018darts, real2019regularized}.
Our work apply meta learning in the context of  optimizing the learning based on heterogeneous experts in building unified user segmentation systems.

\newcommand{\task}{\mathcal{T}}
\newcommand{\loss}{\mathcal{L}}
\newcommand{\inp}{\mathbf{x}}
\newcommand{\learner}{f}
\newcommand{\lossi}{\loss_{\task_i}}

\section{Problem Overview}
As illustrated in \autoref{fig-illustration},
our unified system of user segmentation will ingest items of interests from a variety of domains such as Hosted Content, Mobile, Advertisement and Finance
with a diverse range of knowledge enrichment, resulting in a heterogeneous information network of users and events, and existing segments.
each with their schema, modality and patterns of interconnection.
The first step of our pipeline is to perform data integration to construct a set of \textit{unfolded concept}
where
where different types of interconnections
between entities in the information network
are serialized as an atomic concept
\cite{li2021automl}, following the \textit{Hadoop-MTA} \cite{li2021automl} approach.

Formally, in order to predict a particular segment, let $\mathcal{S}$ be the set of users (i.e. entity in \cite{li2021automl}) that we predict the segment for
and $\mathcal{Y}$ be the set of possible labels.
 We represent the resulting \textit{unfolded concepts} as a real-valued concept vector
$(\vec{c_s})$
for each user $s$,
with index being the list of concept vocabulary $\mathcal{C}$
and value being the intensity of its association to corresponding concepts.

For clarity, we first describe the scenario for learning with homogeneous expert.
Specifically, we assume a particular expert $h_j$
associated with a hypothesis space $H_{j} \subseteq \mathbb{R}^{\mathcal{C}} \to \mathcal{Y}$.
We abstract the algorithm for training the expert and assume an efficient oracle $\theta^{*}_j (\omega; \mathcal{D})$ for obtaining the trained experts
based on a given dataset $\mathcal{D}$
and meta-parameter  $\omega \in \Omega$ that controls how the models are learned, such as model hyperparameters \cite{franceschi2018bilevel}.

:
\vspace{-10pt}
\begin{align} 
&
 \theta^{*}_j (\omega; \mathcal{D})
 \defeq \arg\min_{\theta_j \in \Theta_j} R^{\mathcal{D} }_j(h_{j}(\cdot;\theta_j) )  \nonumber\\
& = \sum_{s \in \mathcal{D} }
\mathbf{L}_j(h_j(\vec{c_s}; \theta_j), \mathbf{y}(s)) \label{eq:oracle}
\end{align}
Here $\theta_j$ is the set of learn-able parameters contained in the parameter space $\Theta_j$,
and $\mathbf{L}_j$ is the loss used for training $h_j$, e.g. the loss function used for back-propagation.

The task of unfolded concept learning task
with \textit{homogeneous} expert will then utilize one such oracle.
It can be stated as follows
\begin{Definition} [Unfolded Concept Learning With Homogeneous Expert] \label{def concept}
Assuming the label function of interest
$\mathbf{y}:\mathcal{S} \to \mathcal{Y}$ mapping each user to a label in $\mathcal{Y}$, a probability density of the entity $\mathbf{q} : \mathcal{S} \to [0,1]$, and a sampled dataset $\mathcal{D}$,
the task is to learn a model $\mathbf{h}_j \in H_{j}$,
that minimize the expected risk
according to a given criterion $\mathbf{L}$

\begin{equation*}
\begin{aligned}
& \underset{\omega}{\text{minimize }}
R(h_{j}(\cdot;\cdot,\omega) )
\defeq \mathbb{E}
_{\mathbf{q}}[\mathbf{L}(h_j(\cdot; \theta^{*}_j (\omega; \mathcal{D})), \mathbf{y}(s))]\\
&  = \int_{\mathcal{S}}
\mathbf{L}(h_j(\vec{c_s}; \theta^{*}_j (\omega; \mathcal{D})), \mathbf{y}(s)
)
\mathbf{q}(s) d s\\
\end{aligned}
\end{equation*}
where $\theta_i \in \Theta_i$ denotes the task specific parameter
and $\omega \in \Omega$ denotes the meta-parameter. 
\end{Definition}

\begin{figure*}[]
\centering
\includegraphics[width=\textwidth]{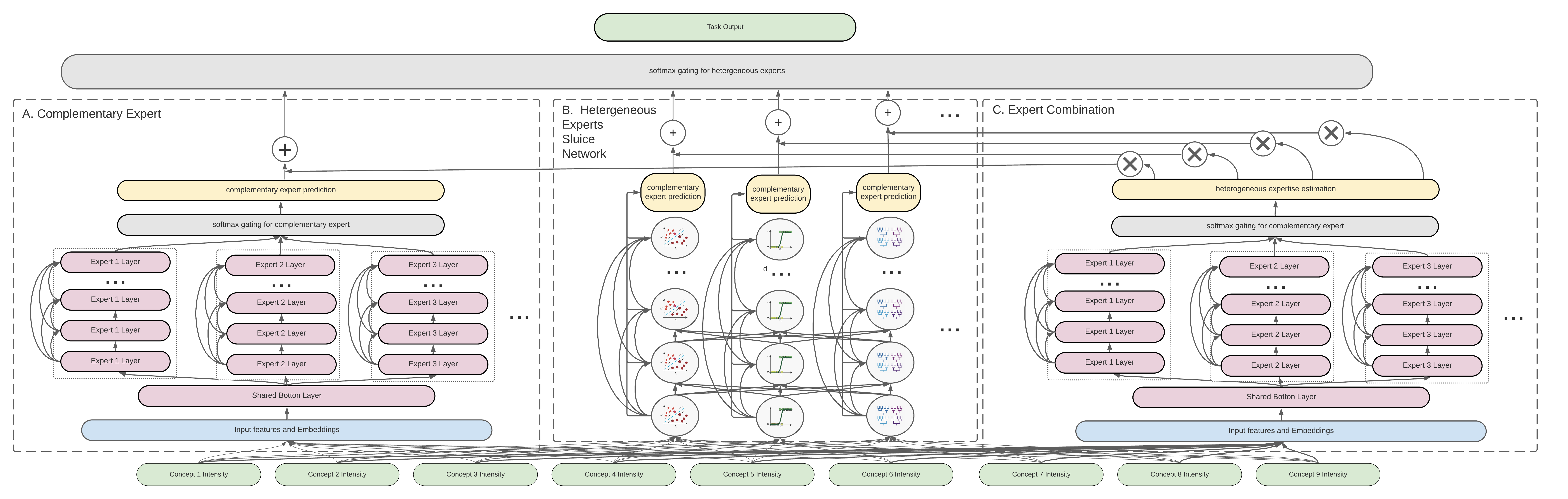}
\caption{Illustration of architecture of the meta-module of \textit{SuperCone}.
Heterogeneous experts are interconnected and reapplied recursively, where all intermediate outputs are allowed to participate in the final prediction, which is gated by the expert combination network.
A neural network for learning the complementary expert are learned in parallel to the heterogeneous experts to allow sharing and complementing the expertise.}
\label{fig-network}
\vspace{-10pt}
\end{figure*}

\vspace{-10pt}

We start the formalization of the user segmentation problem by considering the meta-learning problem in  a more general setting.
Assuming a distribution over tasks $\mathbf{p}_{\mathbb{T}}: \mathbb{T} \to [0,1]$,
we first assume a source (i.e. meta training) dataset of $M$ tasks sampled from $\mathbb{T}$,
each containing a training set
(i.e. support set in meta learning literature \cite{bechtle2021meta}) and
validation set (query set in meta learning literature \cite{bechtle2021meta}) with
 non-overlapping i.i.d. samples
 drawn from instances distribution $\mathbf{q}_j$ of task $\mathbf{T}_j$
 , as
$\mathscr{D}_{source} \defeq \{(\mathcal{D}^{train~(j)}_{source}, \mathcal{D}^{val~(j)}_{source}\}_{j=1}^M$.
Likewise, we assume a target dataset  (i.e. meta test) of $Q$ tasks sampled from $\mathbb{T}$,
each containing a training set
(i.e. support set) and
test set (query set) with
 non-overlapping i.i.d. samples
 drawn from instances distribution $\mathbf{q}_j$ of task $\mathbf{T}_j$
 , as $\mathscr{D}_{target} \defeq \{\mathcal{D}^{train~(j)}_{target}, \mathcal{D}^{test~(j)}_{target}\}_{j=1}^Q$.
The goal is to obtain the ``meta knowledge'' in the form of $\omega$ from $\mathscr{D}_{source}$ which will then be applied to improve downstream  performance in $\mathscr{D}_{target}$,
by fine-tuning on each individual training set at meta-test time.

For the task of learning with heterogeneous experts, however, we focus on the scenario where  source and target set are not separate.
Specifically,
we only require one dataset $\mathcal{D}$
to serve as the source dataset, for meta-training,
and one target dataset, for meta-test.
We assume each of task $j$, $j = 1 \ldots J$, where
the only difference between tasks is the
particular expert $h_j$,
each associated with a hypothesis space $H_{j} \subseteq \mathbb{R}^{\mathcal{C}} \to \mathcal{Y}$, a set of learn-able parameter $\theta_j \in \Theta_j$, and a training oracle
$\theta^{*}_j (\omega; \mathcal{D})$ satisfying \autoref{eq:oracle}. The end goal of meta training, then, is to obtain optimal generalization error on the single test target set.

Formally, we assume
all the available instances will be used for both the source and target set.
 Given a sample of data
$\mathscr{D} \defeq \{\mathcal{D}^{train}, \mathcal{D}^{test}\}$
drawn i.i.d from the instance distribution $\mathbf{q}(s)$.
We will use some or all of the instances from $\mathcal{D}^{train}$ for training the individual experts $h_j(\cdot;\theta_j,\omega)$, i.e.
$(\mathcal{D}^{train~(j)}_{source} \subseteq \mathcal{D}^{train},  \mathcal{D}^{val~(j)}_{source} \subseteq \mathcal{D}^{train}$,
$\mathcal{D}^{train~(j)}_{source} \cap \mathcal{D}^{val~(j)}_{source} = \emptyset $.
Likewise, the dataset used for meta-test consume some or all of the training instances, i.e.,
$\mathcal{D}^{train~(j)}_{target} \subseteq \mathcal{D}^{train}, j = 1\ldots J$.
The final goal is to learn a joint model based on the adapted experts on the target training set,
$\theta^*_j(\omega; (\mathcal{D}^{train~(j)}_{target})$ for $j = 1\ldots J$, denoted as \\
$h(\cdot; \omega, \{h_j(\cdot; \theta^*_j(\omega; \mathcal{D}^{train~(j)}_{target})) \})$,
that achieves the best generalization error.

\vspace{-10pt}

\begin{Definition} [Unfolded Concept Learning With Heterogeneous Experts]
\label{def concept}
Assuming the label function of interest
$\mathbf{y}:\mathcal{S} \to \mathcal{Y}$, a
a sampled dataset $\mathscr{D}$, a set of heterogeneous experts $h_j$ with inner training oracle $\theta^*_j(\omega,\mathcal{D})$ for $j = 1\ldots J$,
the task is to learn a combined model $h$ that minimize a given loss criteria $\mathbf{L} : \mathcal{Y} \times \mathcal{Y} \to \mathbb{R}$
\begin{align}
& \underset{}{\text{minimize }}
R^{\mathcal{D}^{test}}_{\{H_{j}\}^J_{j=1}, \Omega}
\defeq \nonumber\\
& = R^{\mathcal{D}^{test}}(h(\cdot; \omega^*, \{h_j(\cdot; \theta^*_j(\omega^*; \mathcal{D}^{train~(j)}_{target})) \}) ) \nonumber\\
 & =
 \sum_{s \in \mathcal{D}^{test} }
\mathbf{L}((h(\vec{c_s}; \omega^*, \{h_j(\cdot; \theta^*_j(\omega^*; \mathcal{D}^{train~(j)}_{target})) \}) , \mathbf{y}(s))
) \label{eq:con-meta}\\
 \text{s.t. } &\omega^* = \arg\min_{\omega} \mathbf{L}^{meta}(
 \{
 \theta^*_j(\omega, \cdot) | j = 1 \ldots J
 \},
 \omega, \mathcal{D}^{train}
 )
\end{align}

\noindent where
$\mathbf{L}^{meta}$ is a meta loss to be specified by the meta-training procedure, such as the cross entropy error of temporal difference error \cite{finn2017model}.
\end{Definition}

The problem of unfolded concept learning with heterogeneous experts 
extends unfolded concept learning for efficient and scalable distributed AutoML
and retains the representation power. By considering its relation to unfolded concept learning problem  \cite{li2021automl}, we have the following problem complexity results
\begin{Theorem}
The above problem of Unfolded Concept Learning With Heterogeneous Experts is no less difficult than
the
Learning In Heterogeneous Data Problem (Definition 1 in \cite{li2021automl}),
Learning In Relational Database (Definition 2 in \cite{li2021automl}),
Heterogeneous Graph Learning (Definition 3 in \cite{li2021automl}), and
First order Logic Graph Learning (Definition 4 in \cite{li2021automl}).
In fact, there exists efficient linear time reduction from
Learning In Heterogeneous Data Problem,
Learning In Relational Database,
Heterogeneous Graph Learning, and
First order Logic Graph Learning to Unfolded Concept Learning With Heterogeneous Experts problem.
\end{Theorem}

    \begin{figure*}[]
    \centering
    \includegraphics[width=\textwidth]{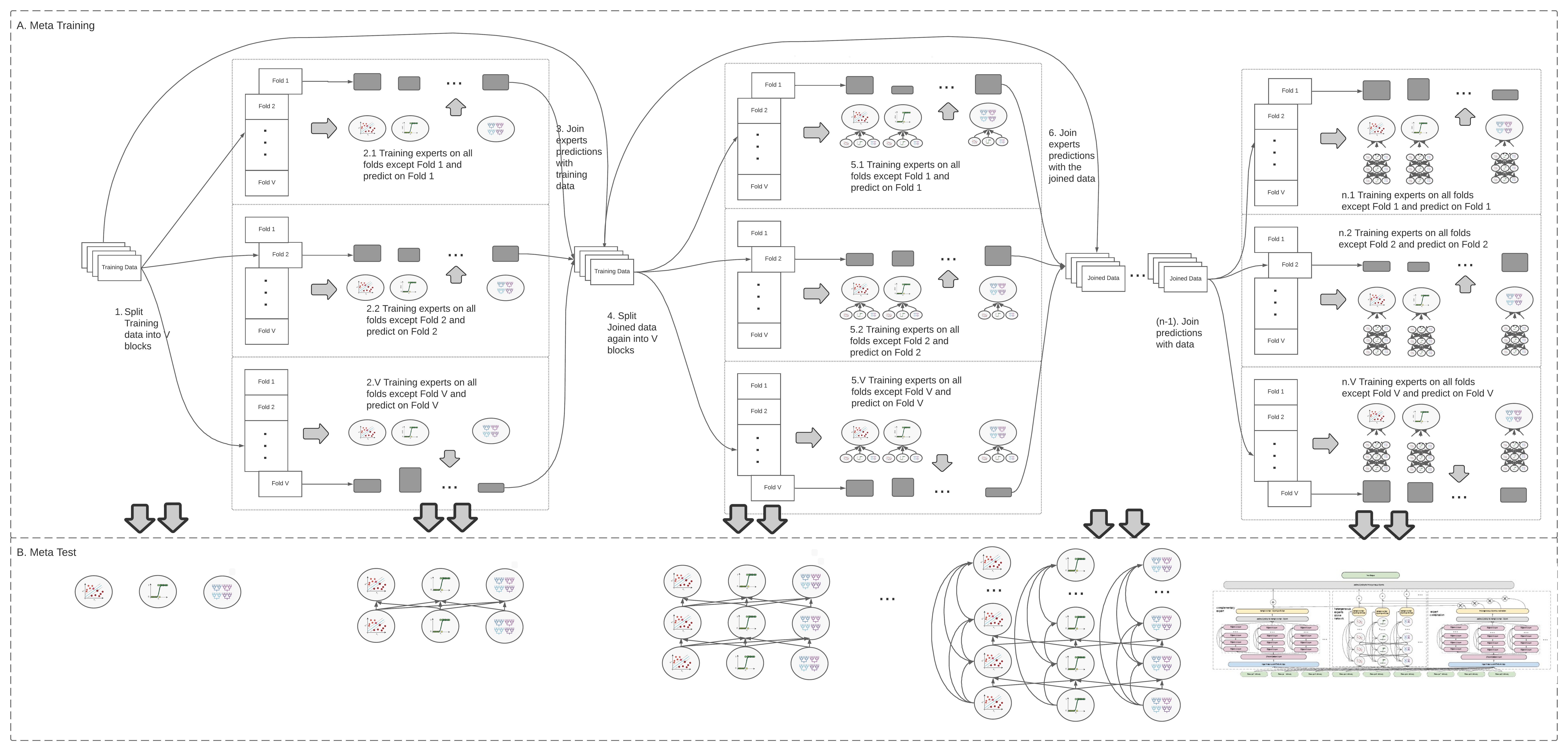}
    \caption{Illustration of meta optimization procedure}
    \vspace{-10pt}
    \label{fig-optimization}
    \end{figure*}

\vspace{-15pt}
\section{Choice of $\Omega$}

    We divide our discussion of our approach into two parts,  the representation of the meta module and the optimization procedure.
    In this section, we focus on the first part of the meta parameter space $\Omega$, for any given choices of $\Theta_j$ of each individual experts $H_j$.

    The solution space induced by meta parameter $\omega$ brings inductive bias to the downstream tasks and affect the efficiency of learning procedure of each task.
    There are several key challenges for the task of model building in critical user segmentation use cases
    \begin{itemize}
        \item \textbf{Task Agnostic Expertise Modeling}
        The choice of $\Omega$
        should allow flexibly modeling over a large variety of tasks types
        and best utilizing the power of experts from $\mathcal{H} = \{H_j | j = 1 \ldots J\}$ in an adaptive without task-specific engineering.

        \item \textbf{Representation Power}
        The choice of $\Omega$ should contain enough representation capacity for inducing  deep representation of data and not
        not limit itself to specific features or classes of functions
        .
        \item \textbf{First Order Influence}
        The influence of meta parameter $\omega$ over the learning mechanism should allow for efficient meta-optimization for performance critical application, and
         not incurring to higher order gradient computation  \cite{finn2017model}  during  the learning of $\omega$.
        \end{itemize}

    Previous approaches mostly fall into the following categories:
    traditional super learning and ensemble learning approaches \cite{polley2010super} are heuristic in nature and fail to meet the second criteria;
    traditional deep learning approaches \cite{zhao2019recommending} and fails the first one by not incorporating the power of heterogeneous experts;
    the majority of the existing meta-learning approaches relies on higher order and bi-level optimization \cite{finn2017model} \cite{franceschi2018bilevel}  and are disqualified by the third criteria.

    To address this, we present the \textit{SuperAug} meta-learning architecture that constructs a large portfolio of augmented experts and learns deep representation for both direct prediction from unfolded concepts and indirect combination of heterogeneous experts,
    while at the same time respecting their individual prediction power and expertise interpret-ability. It consists of the following 3 main components.

    \noindent\textbf{Heterogeneous Experts Sluice Network}
    Upon a given set of heterogeneous experts, we aim to construct a augmented set of experts  $\mathcal{H}_{Aug}$ by
    by enumerating nested combinations
    among the experts.
    Specifically, the space of  experts $\mathcal{H}_{Aug}$ follows the given rules:

    \begin{enumerate}
        \item Any expert model with hypothesis space $H$ belonging to initial experts $\mathcal{H}$ will also belong to $\mathcal{H}_{Aug}$
        \item Any arithmetic combination between an arbitrary number of experts in  $\mathcal{H}_{Aug}$  will also belong to $\mathcal{H}_{Aug}$
        \item Any recursive application of an expert with hypothesis $H$ belong to
         $\mathcal{H}_{Aug}$  over an arbitrary number of outputs from models from $\mathcal{H}_{Aug}$
         will also belong to $\mathcal{H}_{Aug}$
    \end{enumerate}
    The Expert Expansion is implemented in a heterogeneous expert network  in \textit{SuperAug} following the sluice network architecture \cite{ruder2019latent}, with an additional layer by layer skip connections.
    As shown in \autoref{fig-network}, the output at each level of densely connected experts $\sigma(\cdot)$ will be fed to both the immediately next level as input as well as future levels, and the subsequent connected layers henceforth.

    \noindent\textbf{Complementary Expert}
    To further augment the model capacity and obtain deep representation of the data,
    we incorporate a complementary expert module with hypothesis $H_{Comp}$ that allows flexible modulation of information flow while respecting the simplicity of network design. 
    To that end,
    we follow the neural multi-mixture of experts  architecture that learns an ensemble of individual experts in an end to end fashion \cite{jacobs1991adaptive, ma2018modeling}.
    
    Specifically, we divide the neural net into the following:
    the end output module $\mathbf{Tower}$ for producing the output for the particular task;
    the inner expert neural submodules $\mathbf{InnerExpert}_t$, $1\leq t \leq E$;
    and the gating network $\mathbf{Gate}_i$ that projects the input into $\mathbb{R}^E$ directly from the original data representation $\vec{c_s}$.
    The prediction of the final complementary expert that map concept vector representation $\vec{c_s}$ into label space
    $\mathcal{Y}$, $H_{Comp}(\vec{c_s})$,
    can then be expressed as
    \begin{align}
        h_{alt}(\vec{c_s}) & = \mathbf{Tower}( v_s  ) \\
                v_s &  = \sum_t^{E} \Big( \textrm{softmax} (\mathbf{Gate}(\vec{c_s}))_{(t)} \cdot \mathbf{InnerExpert}_t(\vec{c_s}) \Big)
    \end{align}
    Here the intermediate representation $v_s$ is a weighted sum by a shallow network $\mathbf{Gate}_i(\vec{c_s^{meta}})$ after normalizing into unit simplex via $\textrm{softmax}(\cdot)$.
    Each $\mathbf{InnerExpert}_t$, will, in turn, be an ensemble of submodules mapping $\vec{c_s}$ to a fixed-length vector. 
        \begin{align}
        \mathbf{InnerExpert}_t(\vec{c_s}) & = \sum_{i=0}^{L} \mathbf{Depth}_{t,i}(\vec{c_s}) \\
                \mathbf{Depth}_{t,i} & =
        \mathbf{Proj}_{t,i} \Big(\mathbf{Proj}_{t,i-1}
        \Big(\ldots
        (\mathbf{Embed}\Big(\vec{c_s}\Big)
        \ldots
        \Big)
        \Big)
    \end{align}
    Here $\mathbf{Depth}_{t,i}$ denotes an intermediate output for the inner expert $t$ at depth $i$, consisting of projection in the form of $\mathbf{Proj}_{t,i}$, implemented as a linear layer followed by a $relu$ activation.

As illustrated in the alternative expert component in \autoref{fig-network}, an ensemble of neural experts will first be combined to form a deep representation from the concept vector, and further be combined with the rest of heterogeneous experts.

    \noindent\textbf{Expert Combination}
     One distinctive advantage of \textit{SuperAug} over traditional ensemble approaches is the ability to adaptively weigh-in different predictions across experts adaptively.
    To that end, we follow the DARTS meta learning \cite{liu2018darts} design for building the expert combination module.
    Here assuming the experts from $\mathcal{H}_{Aug}$ are arranged as a array of mappings $\{h_1, h_2, \ldots, h_{\mathcal{H}_{Aug}}\}$,
    the combination network component $\mathbf{Comb}(\cdot)$, 
    will map the concept vector $\vec{c_s}$ into a $|\mathcal{H}_{Aug}| + 1$ dimension vector. 
    The final model prediction, $h(\vec{c_s})$, is then produced using another layer of weighted sums over all possible experts
    \begin{align}
    h ( \vec{c}_s ) = \sum_{t \in \{1, 2,\ldots, T\} \bigcup \{Aug\}} \Big( \textrm{softmax} (\mathbf{Comb}(\vec{c_s}))_{(t)} \cdot h_t(\vec{c_s}) \Big)  \label{eq:final}
    \end{align}

\section{meta optimization}
    In this section, we describe the approach for optimizing the meta-parameters $\omega$ , that are agnostic to the heterogeneous experts in $\mathcal{H}$.
    Naive approach that directly treats the original input dataset $\mathscr{D}$ to compute the meta loss $\mathbf{L}^{meta}$ or using it as the support set $(\mathcal{D}^{train~(j)}_{source}$ might lead to "meta-overfit" where the combination network and the added experts from $\mathcal{H}_{Aug}$ falsely rely on overfitted experts.
    In contrast, we propose a principled framework to construct a meta-training set that eliminates the phenomenon and achieves generalization with provable guarantee.
     The high level intuition is to
     extract non-overlapping subset of the data as the support and query set as the source data meta-training to minimize the discrepancy between meta-training and deployment.
     Our optimization method makes no assumption about the heterogeneous experts, including the existence of gradients in its learning process.

     The optimization is shown in \autoref{fig-optimization}, where each level of heterogeneous experts is trained recursively on previous levels with its own meta-training set based on the cross-validation split,
     with the final level corresponding to the \textit{SuperAug} architecture.
     Specifically, we can index heterogeneous experts by the depth it depends on other experts, with $h ^{(k)}_j$ denoting the $j$th expert at $k$th layer, $k = 1, 2,\ldots, K$.
     At each depth, we have a cross validation scheme, $V^{(k)}$ mapping instance $s$ from $\mathcal{D}^{train}$ to a fold among $1, 2, \ldots, V$, the learning proceed by creating higher-order meta training dataset at each $k$th layer, $\mathcal{D}^{train~(k)}$ as
     \begin{align}
\mathcal{D}^{train~(k)} & \defeq \{ (\vec{x^{(k)}_s}, \vec{z^{(k)}_s}|
\vec{x_s} \in \mathcal{D}^{train~(k)}
\} \label{eq:cvtrain1}\\
\vec{z^{(k)}_s}_{(j)} & \defeq
h ^{(k)}_j(\vec{x_s};\theta^*_j(\omega,(\mathcal{D}^{train~(k-1)})^{ \sim s}) ) \label{eq:cvtrain2}
     \end{align}
with $({D}^{train~(k)})^{ \sim s}$ denoting the subset of $({D}^{train~(k)})$ not in the same fold as instance $i$, formally
\begin{align}
    ({D}^{train~(k)})^{ \sim s} \defeq \{V^{(k)}(s) \neq V^{(k)}(s') | \vec{x_{s'}} \in \mathcal{D}^{train~(k)}\} \label{eq:cvtrain3}
\end{align}
And the meta-parameter $\omega$ is trained using the last layer of the constructed meta-training dataset
$\mathcal{D}^{train}_{source} \defeq \mathcal{D}^{train~(K)}$,
with respect to the meta loss defined as follows
    \begin{align}
& \mathbf{L}^{meta}(
 \{
 \theta^*_j(\omega, \cdot) | j = 1 \ldots J
 \},
 \omega, \mathcal{D}^{train}
 ) \defeq   \sum_{\vec{x_s} \in \mathcal{D}^{train}_{source}} \mathbf{L}\Big(h^{train}(\vec{x_s}),
     \mathbf{y}(s)
 \Big) \label{eq:metaloss}
 \end{align}
 with the meta-training time model $h^{train}(\vec{x_s})$ defined by replacing the output of all heterogeneous experts directly by taking all but the first $|\mathcal{C}|$ elements from the input, $\vec{x_s}_{[:|\mathcal{C}|]}$ and feeding the alternative expert and the combination network with the original feature, $\vec{x_s}_{
     [:|\mathcal{C}|]
     }$. Formally,
 \begin{align*}
    h^{train} ( \vec{x} ) &\defeq \sum_{t \in \{1, 2,\ldots, T\} \bigcup \{Aug\}} v_s^t \\
    v_s^t &\defeq \Big(\textrm{softmax} (
     \mathbf{Comb}(\vec{x_s}_{
     [:|\mathcal{C}|]
     })
     \Big)_{(t)}
      \cdot \Big(
     h_{alt}(\vec{x_s}_{[:|\mathcal{C}|]}), \vec{x_s}_{[|\mathcal{C}|:]}
     \Big)
     _{(t)}
\end{align*}
 The learning of the network parameter thus become an end-to-end optimization problem which can be solved using efficient gradient based methods \cite{liu2018darts}.

 Finally, at meta-test time, the source set for each of the heterogeneous experts $h ^{(k)}_j$, $\mathcal{D}^{train~(k,j)}_{target}$ is defined as the $k$-th high order meta training dataset, i.e. $\mathcal{D}^{train~(k,j)}_{target} \defeq  \mathcal{D}^{train~(k)}$
    .
We also have the following results regarding the model's asymptotic and finite sample generalization error over arbitrary heterogeneous expert or the meta learning architecture.

\begin{Theorem}
Assume $\mathcal{Y}$ with bounded cardinality, for any prediction model $\mathbf{y}'$, there exists an parameter space of \textit{SuperAug}  $\Omega$ with the same or less generalization error on instance distribution $\mathbf{q}(s)$  for every instantiation of the data $\mathscr{D}$ in an asymptotic sense.
Moreover, for float-point based implementation of meta-parameter $\omega$, and $K=1$, then its generalization error will converge to 0 or to the best prediction model under a $O(\frac{log n}{n})$ rate.
\end{Theorem}
\noindent\textbf{Proof}
We start with the case of asymptotic generalization error.
Consider an arbitrary prediction model $h_1(\cdot)$ with a learning oracle $\theta^*_1( \mathcal{D})$, we construct the following \textit{SuperAug} architecture with a series of heterogeneous experts including $h_1(\cdot)$,
W.L.O.G. we assume it is the first expert with index 1, since the  \textit{SuperAug} architecture  will further optimize the training time error compared to its input, with probability at least $1 - \delta$ we have
\begin{align}
& \int_{s \in \mathcal{S}}
\mathbf{L}((
h(\vec{c_s}; \omega^*, \{h_j(\cdot; \theta^*_j(
\omega^*; \mathcal{D}^{train~(j)}_{target}
)) \})
, \mathbf{y}(s))
) \mathbf{q}(s) \nonumber\\
\leq & \sum_{s \in \mathcal{D}^{train}_{source} }
\mathbf{L}((h(\vec{c_s}; \omega^*, \{h_j(\cdot; \theta^*_j(\omega^*; \mathcal{D}^{train~(j)}_{target})) \}) , \mathbf{y}(s))
)   + \nonumber\\ & \mathcal{O}
\bigg(
\sqrt{
 \frac{C_1 log |\mathcal{D}^{train}| + C_2 + log 1/\delta}{|\mathcal{D}^{train}|}
)
}
\bigg) \nonumber\\
\leq & \int_{s \in \mathcal{S}}
\mathbf{L}(h_1(
\cdot; \theta^*_1( \mathcal{D}^{train})
), \mathbf{y}(s))
 \mathbf{q}(s)  + \nonumber\\ &\mathcal{O}
\bigg(
\sqrt{
 \frac{C_1 log |\mathcal{D}^{train}| + C_2 + log 1/\delta}{|\mathcal{D}^{train}|}
)
}
\bigg)
\nonumber\\
\end{align}

where the first and second in-equality is established with \cite{vapnik1999nature} and $C_1$, $C_2$ are fixed constant.
For the second part of the theorem,
again consider an arbitrary prediction model $h_1(\cdot)$, we construct 1 level  \textit{SuperAug} architecture with a series of heterogeneous experts including $h_1(\cdot)$ as the first expert with index 1, along with a series of experts that output the original feature $\vec{x_s}$ into the expert combination $\omega$. If we denote $h^*(\cdot)$ as the expected risk minimizer and $d(h, h^*) \defeq E_{s \sim \mathbf{q}(s)}(
\mathbf{L}(
h(\vec{x_s}), \mathbf{y}(s)
) -
\mathbf{L}(
h^*(\vec{x_s}), \mathbf{y}(s)
)
)$ be the expected performance of a model $h$,
by leveraging
the
 results in Equation 2 in \cite{van2007super},
  from which the convergence results will follow from the fact that
  for every $\delta > 0$ there exists a constant $C$ that
\begin{align}
    & \frac{1}{V} \sum_{v=1}^V E d(h(\vec{c_s}; \omega^*, \{h_j(\cdot; \theta^*_j(
\omega^*; \{\vec{x_s} \in \mathcal{D}^{train}, V^{(0)}(s) = v\}
)) \}), h^*) \nonumber\\
\leq & (1+\delta) E \min_{\omega \in \Omega} \frac{1}{V}  \sum_{v=1}^V  d(h(\vec{c_s}; \omega, \{h_j(\cdot; \theta^*_j(
\omega; \{\vec{x_s} \in \mathcal{D}^{train}, V^{(0)}(s) = v\}
)) \}), h^*) \nonumber\\ &
+ C \frac{V log |\mathcal{D}^{train}|}{|\mathcal{D}^{train}|}
\end{align}
\vspace{-10pt}

\begin{table*}[]
\centering
\resizebox{\textwidth}{!}{%
\begin{tabular}{|c|c|c|cc|cc|cc|cc|cc|cc|}
\hline
\rowcolor[HTML]{C0C0C0} 
 &
   &
  WDL &
  \multicolumn{2}{c|}{\cellcolor[HTML]{C0C0C0}PLE} &
  \multicolumn{2}{c|}{\cellcolor[HTML]{C0C0C0}MMOE} &
  \multicolumn{2}{c|}{\cellcolor[HTML]{C0C0C0}ESSM} &
  \multicolumn{2}{c|}{\cellcolor[HTML]{C0C0C0}DCNMix} &
  \multicolumn{2}{c|}{\cellcolor[HTML]{C0C0C0}DCN} &
  \multicolumn{2}{c|}{\cellcolor[HTML]{C0C0C0}SuperCone} \\ \hline
\rowcolor[HTML]{9B9B9B} 
{\color[HTML]{000000} } &
  {\color[HTML]{000000} } &
  {\color[HTML]{000000} Absolute} &
  \multicolumn{1}{c|}{\cellcolor[HTML]{9B9B9B}{\color[HTML]{000000} Absolute}} &
  {\color[HTML]{000000} Relative} &
  \multicolumn{1}{c|}{\cellcolor[HTML]{9B9B9B}{\color[HTML]{000000} Absolute}} &
  {\color[HTML]{000000} Relative} &
  \multicolumn{1}{c|}{\cellcolor[HTML]{9B9B9B}{\color[HTML]{000000} Absolute}} &
  {\color[HTML]{000000} Relative} &
  \multicolumn{1}{c|}{\cellcolor[HTML]{9B9B9B}{\color[HTML]{000000} Absolute}} &
  {\color[HTML]{000000} Relative} &
  \multicolumn{1}{c|}{\cellcolor[HTML]{9B9B9B}{\color[HTML]{000000} Absolute}} &
  Relative &
  \multicolumn{1}{c|}{\cellcolor[HTML]{9B9B9B}Absolute} &
  Relative \\ \hline\hline
 &
  Accuracy &
  0.8227 &
  \multicolumn{1}{c|}{0.8405} &
  +2.16\% &
  \multicolumn{1}{c|}{0.8413} &
  +2.26\% &
  \multicolumn{1}{c|}{0.8148} &
  -0.96\% &
  \multicolumn{1}{c|}{0.7985} &
  -2.94\% &
  \multicolumn{1}{c|}{0.8071} &
  -1.90\% &
  \multicolumn{1}{c|}{0.8491} &
  +3.21\% \\ \cline{2-15} 
 &
  AUC &
  0.8687 &
  \multicolumn{1}{c|}{0.8913} &
  +2.60\% &
  \multicolumn{1}{c|}{0.8938} &
  +2.89\% &
  \multicolumn{1}{c|}{0.8774} &
  +1.00\% &
  \multicolumn{1}{c|}{0.8433} &
  -2.92\% &
  \multicolumn{1}{c|}{0.8646} &
  -0.47\% &
  \multicolumn{1}{c|}{0.905} &
  +4.18\% \\ \cline{2-15} 
 &
  F1 &
  0.5122 &
  \multicolumn{1}{c|}{0.6165} &
  +20.36\% &
  \multicolumn{1}{c|}{0.6416} &
  +25.26\% &
  \multicolumn{1}{c|}{0.4289} &
  -16.26\% &
  \multicolumn{1}{c|}{0.3536} &
  -30.96\% &
  \multicolumn{1}{c|}{0.3957} &
  -22.75\% &
  \multicolumn{1}{c|}{0.6658} &
  +29.99\% \\ \cline{2-15} 
 &
  Kappa &
  0.4156 &
  \multicolumn{1}{c|}{0.5181} &
  +24.66\% &
  \multicolumn{1}{c|}{0.5403} &
  +30.00\% &
  \multicolumn{1}{c|}{0.3448} &
  -17.04\% &
  \multicolumn{1}{c|}{0.2701} &
  -35.01\% &
  \multicolumn{1}{c|}{0.3111} &
  -25.14\% &
  \multicolumn{1}{c|}{0.5686} &
  +36.81\% \\ \cline{2-15} 
 &
  Log loss &
  6.1224 &
  \multicolumn{1}{c|}{5.5094} &
  -10.01\% &
  \multicolumn{1}{c|}{5.4797} &
  -10.50\% &
  \multicolumn{1}{c|}{6.3961} &
  +4.47\% &
  \multicolumn{1}{c|}{6.9583} &
  +13.65\% &
  \multicolumn{1}{c|}{6.6613} &
  +8.80\% &
  \multicolumn{1}{c|}{5.212} &
  -14.87\% \\ \cline{2-15} 
\multirow{-6}{*}{a9a} &
  Overall &
   &
  \multicolumn{1}{c|}{} &
  +59.80\% &
  \multicolumn{1}{c|}{} &
  +70.92\% &
  \multicolumn{1}{c|}{} &
  -37.73\% &
  \multicolumn{1}{c|}{} &
  -85.49\% &
  \multicolumn{1}{c|}{} &
  -59.06\% &
  \multicolumn{1}{c|}{} &
  +89.06\% \\ \hline\hline
\cellcolor[HTML]{FFFFFF}{\color[HTML]{222222} } &
  Accuracy &
  0.505 &
  \multicolumn{1}{c|}{0.5033} &
  -0.34\% &
  \multicolumn{1}{c|}{59.00\%} &
  +16.83\% &
  \multicolumn{1}{c|}{0.545} &
  +7.92\% &
  \multicolumn{1}{c|}{0.5950} &
  +17.82\% &
  \multicolumn{1}{c|}{0.5017} &
  -0.65\% &
  \multicolumn{1}{c|}{0.8450} &
  +67.33\% \\ \cline{2-15} 
\cellcolor[HTML]{FFFFFF}{\color[HTML]{222222} } &
  AUC &
  0.5162 &
  \multicolumn{1}{c|}{0.5083} &
  -1.53\% &
  \multicolumn{1}{c|}{62.34\%} &
  +20.77\% &
  \multicolumn{1}{c|}{0.5627} &
  +9.01\% &
  \multicolumn{1}{c|}{0.6101} &
  +18.19\% &
  \multicolumn{1}{c|}{0.5017} &
  -2.81\% &
  \multicolumn{1}{c|}{0.9070} &
  +75.71\% \\ \cline{2-15} 
\cellcolor[HTML]{FFFFFF}{\color[HTML]{222222} } &
  F1 &
  0.5139 &
  \multicolumn{1}{c|}{0.6005} &
  +16.85\% &
  \multicolumn{1}{c|}{59.00\%} &
  +14.81\% &
  \multicolumn{1}{c|}{0.5269} &
  +2.53\% &
  \multicolumn{1}{c|}{0.5744} &
  +11.77\% &
  \multicolumn{1}{c|}{0.6659} &
  +29.58\% &
  \multicolumn{1}{c|}{0.8453} &
  +64.49\% \\ \cline{2-15} 
\cellcolor[HTML]{FFFFFF}{\color[HTML]{222222} } &
  Kappa &
  0.01 &
  \multicolumn{1}{c|}{0.0067} &
  -33.00\% &
  \multicolumn{1}{c|}{18.00\%} &
  +1700.00\% &
  \multicolumn{1}{c|}{0.09} &
  +800.00\% &
  \multicolumn{1}{c|}{0.1900} &
  +1800.00\% &
  \multicolumn{1}{c|}{0.0033} &
  -67.00\% &
  \multicolumn{1}{c|}{0.6900} &
  +6800.00\% \\ \cline{2-15} 
\cellcolor[HTML]{FFFFFF}{\color[HTML]{222222} } &
  Log loss &
  17.0969 &
  \multicolumn{1}{c|}{17.1546} &
  +0.34\% &
  \multicolumn{1}{c|}{1416.11\%} &
  -17.17\% &
  \multicolumn{1}{c|}{15.7153} &
  -8.08\% &
  \multicolumn{1}{c|}{13.9883} &
  -18.18\% &
  \multicolumn{1}{c|}{17.2122} &
  +0.67\% &
  \multicolumn{1}{c|}{5.3536} &
  -68.69\% \\ \cline{2-15} 
\multirow{-6}{*}{\cellcolor[HTML]{FFFFFF}{\color[HTML]{222222} madelon}} &
  Overall &
   &
  \multicolumn{1}{c|}{} &
  -18.35\% &
  \multicolumn{1}{c|}{} &
  +1769.58\% &
  \multicolumn{1}{c|}{} &
  +827.54\% &
  \multicolumn{1}{c|}{} &
  +1865.97\% &
  \multicolumn{1}{c|}{} &
  -41.56\% &
  \multicolumn{1}{c|}{} &
  +7076.21\% \\ \hline
\end{tabular}%
}
\caption{Performance evaluation on the public benchmark datasets of \texttt{a9a} and \texttt{madelon} over the metric:  AUC, Accuracy, F1 score, Kappa Cohen Score (Kappa), Log loss against the ground truth.
Both the absolute value and relative value compared to \textit{WDL} baseline are reported.
}
\end{table*}

\begin{figure*}[]
\centering
\vspace{-10pt}
\includegraphics[width=.9\textwidth]{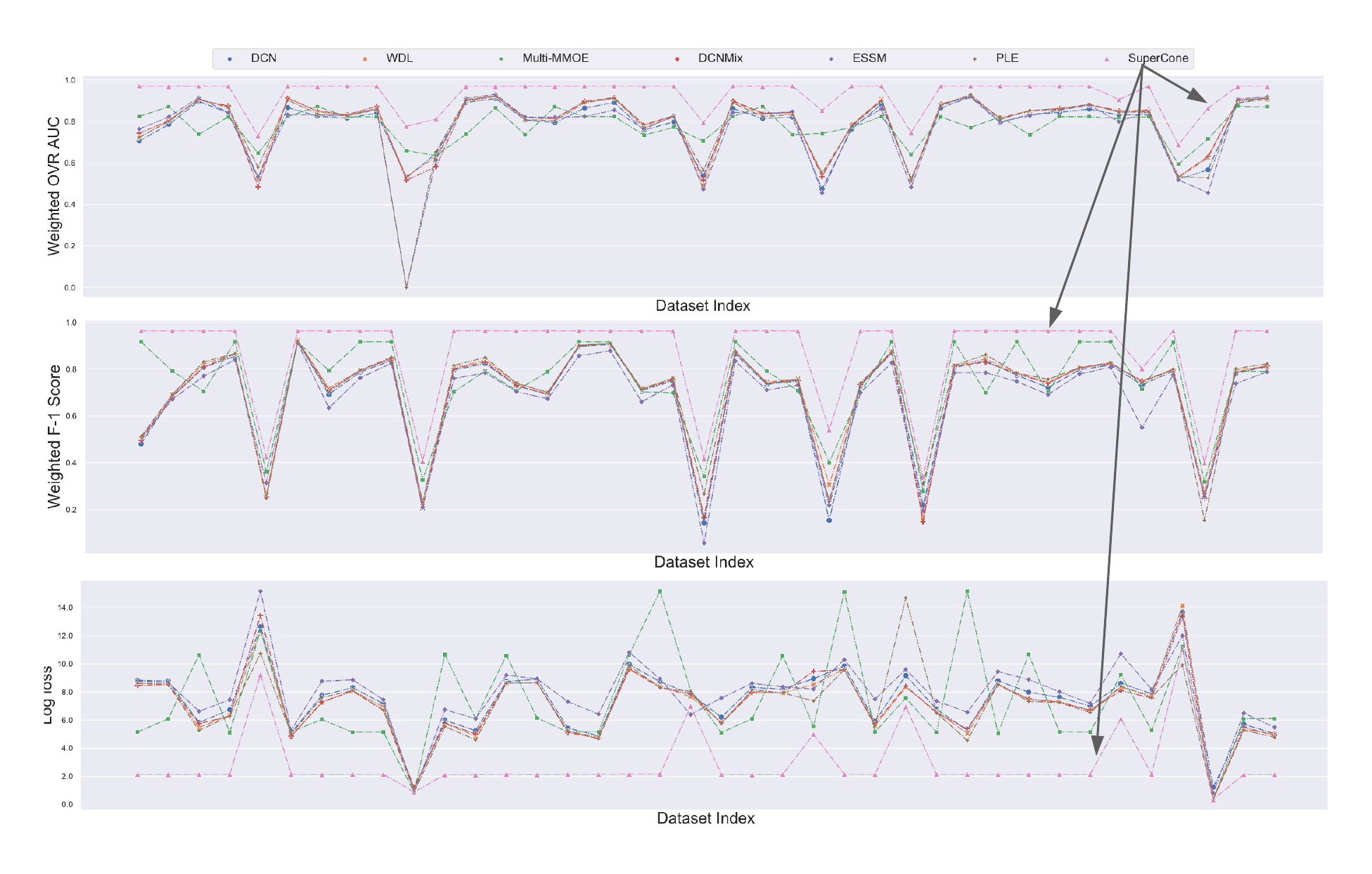}
\vspace{-10pt}
\caption{Performance comparisons of weighted one-versus-result ROC-AUC, weighted F-1 Score, Log loss across the 39 different types of prediction tasks}
\label{fig-AUC}
\vspace{-10pt}
\end{figure*}

\begin{figure*}[]
\centering
\vspace{-10pt}
\includegraphics[width=\textwidth]{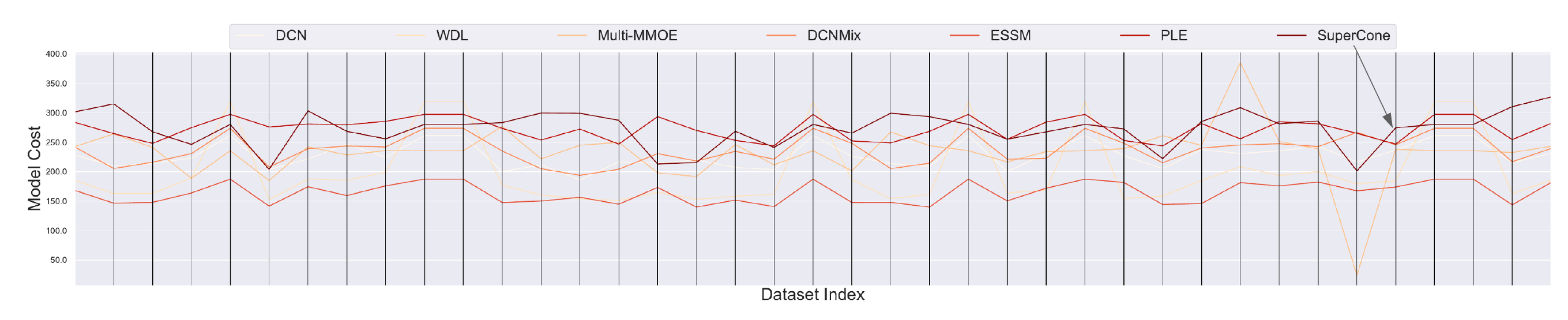}
\caption{Model cost measured in microseconds over the 39 production user segmentation tasks}
\label{fig-inference cost}
\vspace{-10pt}
\end{figure*}

\begin{figure}[]
\centering
\vspace{-10pt}
\includegraphics[width=.5\textwidth]{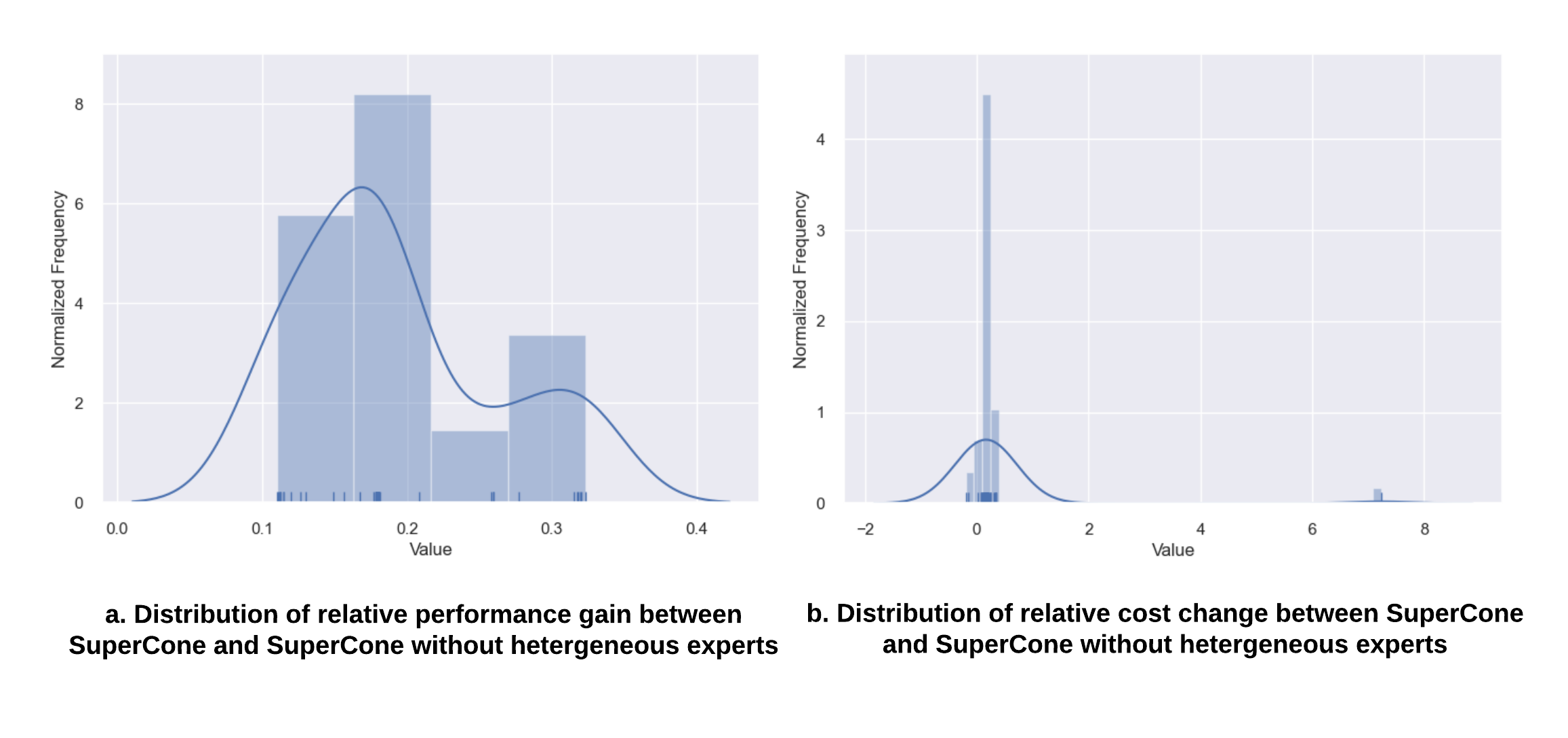}
\vspace{-10pt}
\caption{
Distribution of change in model performance and model cost across \textit{SuperCone} variants.
}
\label{fig-qualitative}
\vspace{-10pt}
\end{figure}

\begin{figure}[]
\centering
\vspace{-10pt}
\includegraphics[width=.5\textwidth]{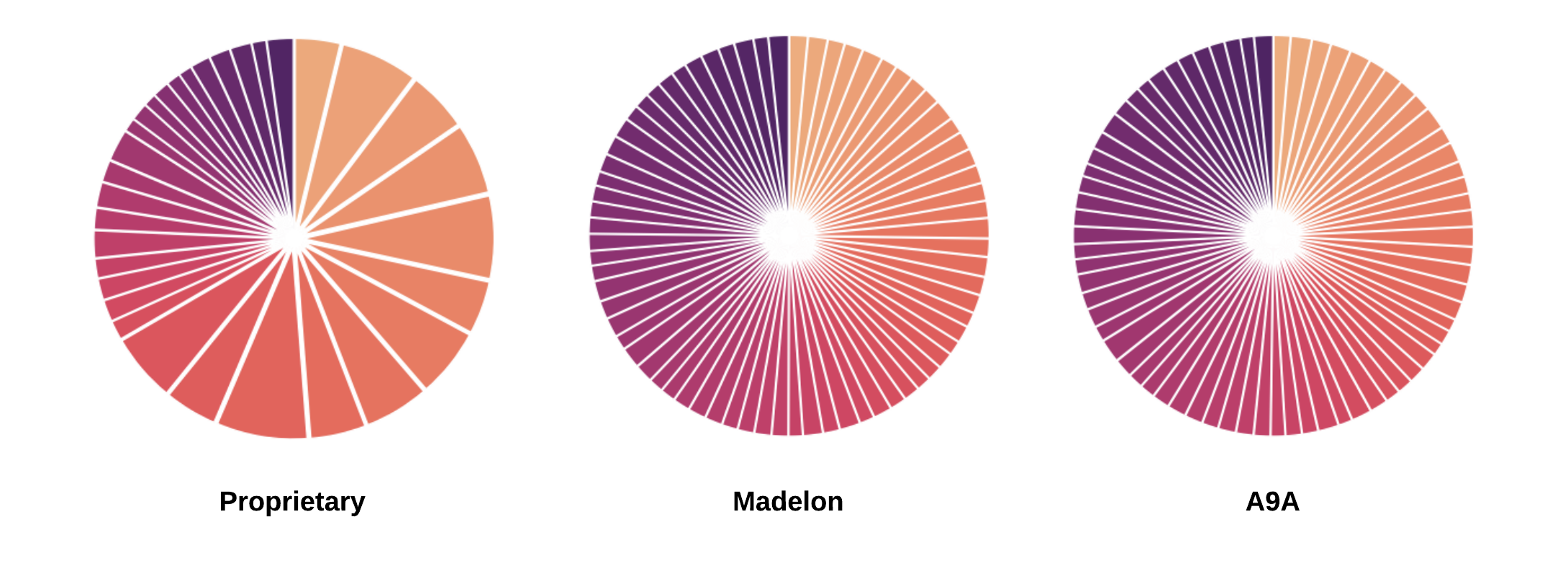}
\vspace{-10pt}
\caption{Experts attention learned by \textit{SuperCone} averages across datasets}
\label{fig-Interpret-ability study on task relation learning}
\vspace{-10pt}
\end{figure}

\section{EXPERIMENT}
In this section, we present a series of experiments centered around the following research questions:
\begin{itemize}[leftmargin=2.5em]
    \item[\bf RQ1] How do  alternative methods compare to SuperCone according to core performance metrics used for production?
    \item[\bf RQ2] How do the settings and individual components of SuperCone affect its quality?
    \item[\bf RQ3] How does the approach of SuperCone compares with other methods when applied to public structured data learning tasks?
    \item[\bf RQ4] Is the approach of SuperCone reliable when applied to different tasks of different types  and domains and interpretable to human inspection?
    \item[\bf RQ5] How does SuperCone perform under resource constrained scenario and balance between the performance and computation cost?
    \item[\bf RQ6] How does SuperCone compared against alternatives in production environment for key end goals?

\end{itemize}

\noindent\textbf{Data-set}
We used both proprietary and public datasets.
For the former, we collected and compiles a total of 39 different user segmentation tasks involving discretized range prediction, multi-class classification prediction as well as binary classification prediction from production.
It is constructed by associating
users with interest taxonomy including
YCT, OIC \cite{zhou2016predicting},
 as well as open-domain knowledge base including Wikipedia and Price-Grabber.
 The dataset contains 100K dimensional unfolded vector per instance, with a total of 100K instances.
 Each of the 39 dataset is split into 3 folds, with 2/3 of them belonging to the support set and remainder belonging to the hold-out query/test set.

We also compare our approaches over several public benchmark dataset.
Specifically, we use the
 \madelon \cite{guyon2004result}
 and
 \ana \cite{platt1998sequential}.
 \madelon contains 2,000 training samples, 600 test samples with 500 features per sample. \ana contains 32,561 training samples, 16,281 test samples with 123 features per sample, respectively.

\noindent\textbf{Methods Comparison}
We implement \textit{SuperCone} in two variants.
The first variant is a homogeneous neural network version that predicts the outcome with purely the neural alternative expert $H_{Comp}$ and the expert combination architecture following a multi-gated neural mixture of expert (\textit{MMOE}) \cite{zhao2019recommending} architecture,
where each one is by itself constructed recursively with an \textit{MMOE},
with the inner MMOE for $H_{Comp}$ having 3 experts 3 layer of densely connected residual connection as shown in Equation 7 with a width of 32, and gate network having 2 layers of densely connected residual connection shown in Equation 7 with width 32,
and the inner MMOE for the combination network having 3 layers of densely connected residual connection as shown in Equation 7 with a width of 32. We denote this the \textit{Multi-MMOE}.
We then use the exact same network architecture and combine it with heterogeneous expert set with $|\mathcal{H}_{Sug}| = 70$ and $K=2$  for public benchmark and the production supported $|\mathcal{H}_{Sug}| = 31$ and $K=1$ experts for the proprietary datasets,
including 11 hyperparameter-tuned gradient boosting models under various implementation trained on GPU accelerators.
The learning rate is tuned using an exponent search and set as 1e-4 with epoch of 30.
The setting is applied to all datasets.

In addition, we implement the following baseline approaches

\begin{itemize}

\item \textbf{PLE} implements the Progressive Layered Extraction method \cite{tang2020progressive} using shared expert count as 1 and  specific expert as 2 , with expert layer width as 256, 256, gate layer width as 16, 16, and tower layer depth as 32, 32.
\item  \textbf{WDL} implements wide and deep learning  \cite{cheng2016wide} with the deep network  layer width tuned as 8 for \madelon and 256, 128, 64 tuned for the rest datasets.
\item  \textbf{ESSM}  implements Entire Space Multi-Task Model \cite{ma2018entire} with CTR component and CVR component each with 
layer width as 512, 512.
\item  \textbf{DCN} implements Deep \& Cross Network \cite{wang2017deep} with layer width as 384, 128, 64, cross count as 2 and cross dimension as 100.
\item  \textbf{DCNMix}  implements Cost-Effective Mixture of Low-Rank DCN \cite{wang2021dcn} with  per layer experts count as 4 and width as 256, 128, 64, cross count as 2 and cross dimension as 100 with a rank of 32.
\end{itemize}

All online adaptation and single task learning was performed with 30 epochs of Adam optimization with a tuned learning rate between 1e-6 and 1e-5 depends on the dataset. The rest settings default to the implementation reported in the  original paper.

\noindent\textbf{Core Performance Evaluation [RQ1, RQ4] }
First and foremost,
we compare the performance of various candidate approaches
over the 39 production user segmentation tasks,
and score their performance using the weighted
one-versus-all ROC-AUC (Weighted OVR AUC) that applies to range-prediction, multi-class prediction and binary prediction,
as well as the weighted F1 score and cross-entropy log loss that also applies to the different
types of prediction tasks simultaneously.
As shown in \autoref{fig-AUC},
\textit{SuperCone} that is implemented agnostic to tasks does not suffer from
overfitting or meta-overfitting, and is able to consistently outperform benchmarks and achieves close to 100 \% F1-score and ROC-AUC without tuning.
With other strong baselines including \textit{Multi MMOE}, \textit{PLE} and \textit{ESSM}.

\noindent\textbf{Public Benchmark Evaluation  [RQ3,RQ4]}
We further evaluate the applicability of \textit{SuperCone} on public structured dataset aganist the best performing version of baselines, where
the Multi MMOE methods degrades to MMOE architecture \cite{zhao2018deep}.
Table 1 reports the absolute value of various performance metrics
including Accuracy, ROC-AUC (AUC), Cohen-Kappa Score (Kappa), F1 Score (F1) and Log loss,
as well as its relevant change compared to baseline \textit{WDL} of
and the aggregated the overall change across metrics.
Again, \textit{SuperCone} without task specific tuning is able to achieve consistent performance, improving on competitive baseline by a significant margin.

\noindent\textbf{Computation Cost [RQ5] }
We then study the computation cost of various approaches that are of critical important for
cost and latency sensitive production system.
Specifically, we measure the computational cost in a per distributed-executor node setting,
where shard-ed dataset are sent to local node and processed sequentially.
\autoref{fig-inference cost} shows the number of microsecond to process a single data instance, where
 \textit{SuperCone} requires similar cost because heterogeneous expert outputs only amounts to small portion of the feature sets,
 and thus achieving a better tradeoff point between performance and cost.

\noindent\textbf{ Ablation Analysis [RQ2]}
We investigate the impact of meta training over heterogeneous experts (see Algorithm 1) by comparing the distribution of
performance gain and cost change in terms of Weighted OVR ROC-AUC over
from \textit{SuperCone} and its ablation version without heterogeneous experts, i.e. the \textit{Multi MMOE} approach.
The left figure of \autoref{fig-qualitative} shows the distribution of relative gain in performance
while the right figure of \autoref{fig-qualitative} shows the distribution of relative cost, aggregated over the 39 production prediction tasks.
We can observe that \textit{SuperCone}
achieve a significant improvement over the already competitive
ablation version with cost  distributed closely around zero in a highly symmetric fashion.

\noindent\textbf{Interpret-ability Study [RQ2,RQ3]}
We next investigate the interpret-ability of \textit{SuperCone}
by visualizing the meta-learned expert attention average across instances and datasets
for the domain of proprietary user segmentation tasks,
\madelon, and \ana.
Specifically, for each dataset,
we extract the instantiated combination network output $\textrm{softmax} (\mathbf{Comb}(\vec{c_s}))_{(t)}$ as shown in Equation 8,
for the $|\mathcal{H}_{Aug}| + 1$ experts, with neural alternative methods followed by the heterogeneous experts.
As shown in \autoref{fig-Interpret-ability study on task relation learning},
expert attention displays an even distribution across the multiple experts,
with the proprietary domain more biased towards models  with GPU accelerator and scale to dataset with much larger instances count and more features.

\begin{table}
\vspace{-10px}

\label{tab:age segment}
\setlength\tabcolsep{1pt} 

\resizebox{.5\textwidth}{!}{
\centering
\begin{tabular}{c|c|c|c|c|c}
\rowcolor[HTML]{C0C0C0}
\textbf{} & \textbf{Accuracy} & \textbf{Recall} & \textbf{Precision} & \textbf{Weighted F1} & \textbf{Cohen Kappa} \\ \hline
Previous Production & 0.20   & 0.23  & 0.28          & 0.21   & 0.08   \\ \hline
SuperCone          & 0.42   & 0.42 & 0.42 & 0.41   & 0.27   \\ \hline
Lift                & +112.54\% & +78.47\% & +47.39\%         & +89.85\% & +249.61\%
\end{tabular}

}

\caption{Comparison of \textit{SuperCone} with the previous production system on range prediction use cases}
\vspace{-10px}
\end{table}
\noindent\textbf{Online evaluation [RQ6,RQ4]}
The \textit{SuperCone} is rolled out to production targeting use cases
in internal Hadoop based deployment system and evaluation in key range predication tasks
Table 2 shows the performance comparison between \textit{SuperCone} and previous production system, which shows that
the meta-training paradigm generalizes well to the new incoming data and compares favorably in the practical setting.

\section{Conclusion}

In this work, we present \textit{SuperCone} as our  solution for user segmentation system that is able to handle
task heterogeneity,
long-tailness and low data availability, by integrating heterogeneous experts and combining them in the end to end fashion, following a principled meta-learning approach.
Extensive evaluation on 39 user segmentation tasks and public benchmarks datasets demonstrate the reliability and superior performance of  \textit{SuperCone} over state-of-the-art recommendation and ranking approaches in key production use cases.

One particular interesting directions for future research is to extend the \textit{SuperCone} paradigm for wider range of modality, business domain and use cases for extended economic and societal impact. 
Another promising direction is to build universal representation and better integration with common knowledge base towards commonsense AI.

\balance

\balance

{\footnotesize
\bibliographystyle{IEEEtran}
\bibliography{_main}
}

\newpage
\appendix
\begin{figure*}[]
\centering

\includegraphics[width=\textwidth]{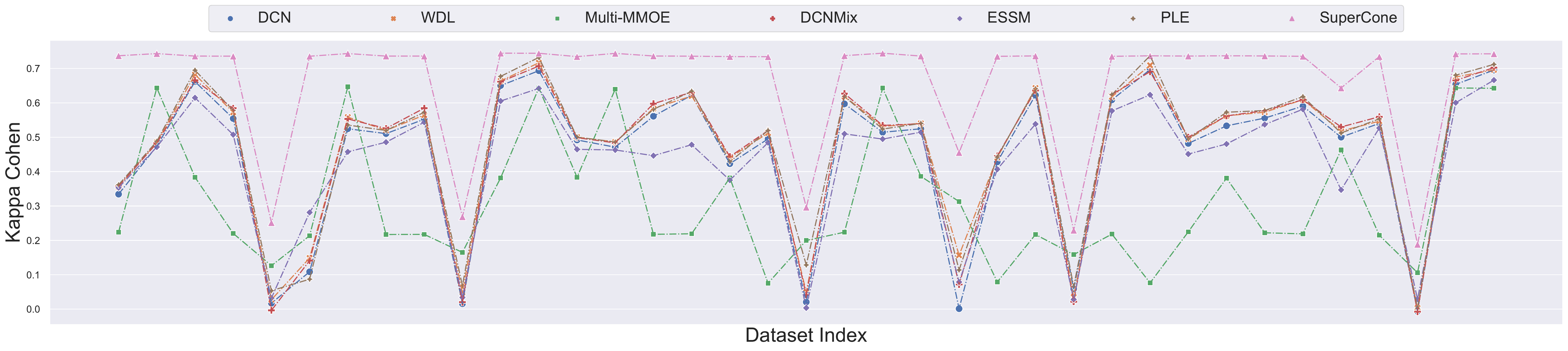}
\caption{Core performance comparison as measured by Kappa Cohen score
over all predictive segment tasks}
\label{fig-Kappa core}
\end{figure*}

\begin{figure*}[]
\centering

\includegraphics[width=\textwidth]{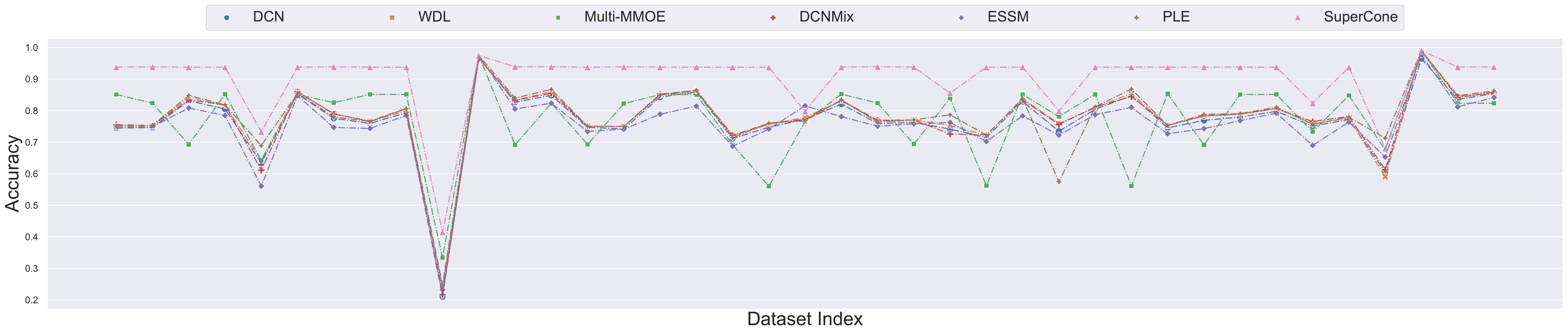}
\caption{Core performance comparison as measured by accuracy over all predictive segment tasks}
\label{fig-Acc core}
\end{figure*}

\section{Details on meta optimization}

    We present the \textit{SuperAug} Algorithm in Algorithm \autoref{alg:maml} that operates on the unfolded concepts.
    It constructs the meta-training set for experts from level 1 to level $K$ in a bottom-up progressive fashion following the cross validation scheme (line 2-7),
    by using the $K$th layer of the meta-data-set for end to end training of meta parameters (line 8), the meta-testing time model can be obtained by adapting on the support set which covers every individual user in the training data as the (line 9-13) and combine them according to original Model architecture (line 14).

The above algorithm for $O(K \cdot J \cdot \frac{n_{experts})}{n_{meta}} + 1$ compared to vanilla differentiable architecture training with 
$\frac{n_{experts})}{n_{meta}}$ being the ratio of average training cost between one single heterogeneous experts and the differentiable architecture.

\section{Details on the \textit{SuperCone} implementation}
\textit{SuperCone} is implemented using the exact same hyper-parameter and optimization setting as the Multi MMOE model,  
together with recursively constructed heterogeneous experts (line 2-7 in Algorithm 1). 
For public benchmark, we use a 
expert set with $|\mathcal{H}_{Sug}| = 70$ and $K=2$ 
, including 14 gradient boosting tree variants, 
1 separately trained relu neural network variant,
8 bagging tree variants, 
7 generalized linear model variants, 
1 Bayesian graphical model variant,
1 nearest neighbor variant,
1 Adaboost variant and 2 SVM variant with model implementation choice set using cross validation in training set .
We use an expert set with $|\mathcal{H}_{Sug}| = 31$ and $K=1$  that is supported by deployment environment,
including 11 hyperparameter-tuned gradient boosting tree models trained on GPU accelerators,  
8 bagging tree variants, 
7 generalized linear model variants, 
1 Bayesian graphical model variant,
1 nearest neighbor variant,
1 Adaboost variant, 
and 2 SVM variant with model implementation choice tuned set cross validation.
The heterogeneous experts are trained on a subset of the corresponding support dataset (see Figure 4) within a time budget of 30 minutes.
These same setting is applied to all datasets in the corresponding domain.

\section{Details on Core Performance Evaluation}
We compare the performance various candidate approaches 
over the 39 production predictive segment tasks
\autoref{fig-Kappa core} and \autoref{fig-Acc core} shows the Kappa Cohen score and accuracy for all tasks, forther demonstrating taht 
\textit{SuperCone} is able to achieve consistently high performance using simple parameter configurations.

\section{Details on the Public benchmark evaluation}
\begin{figure}[]
\centering
\includegraphics[width=.5\textwidth]{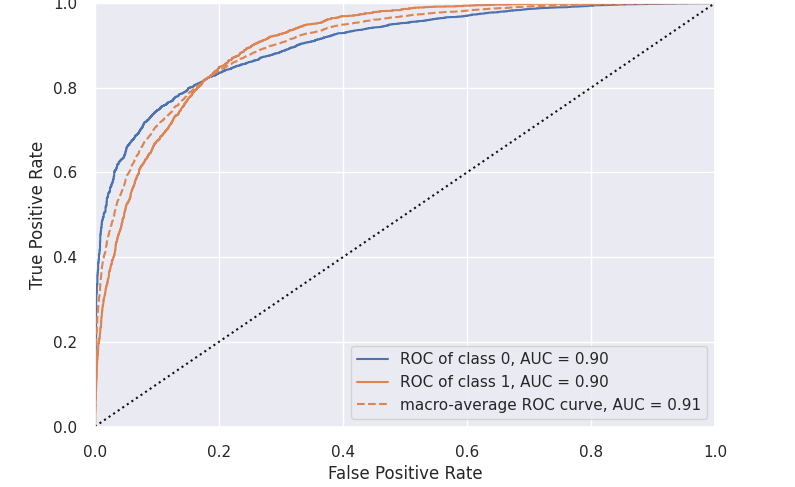}
\caption{Receiver Operating Curve of \textit{SuperCone}
on public benchmark data \texttt{a9a} for both classes
}
\label{fig-a9a auc}
\vspace{-10pt}
\end{figure}

\begin{figure}[]
\centering
\includegraphics[width=.5\textwidth]{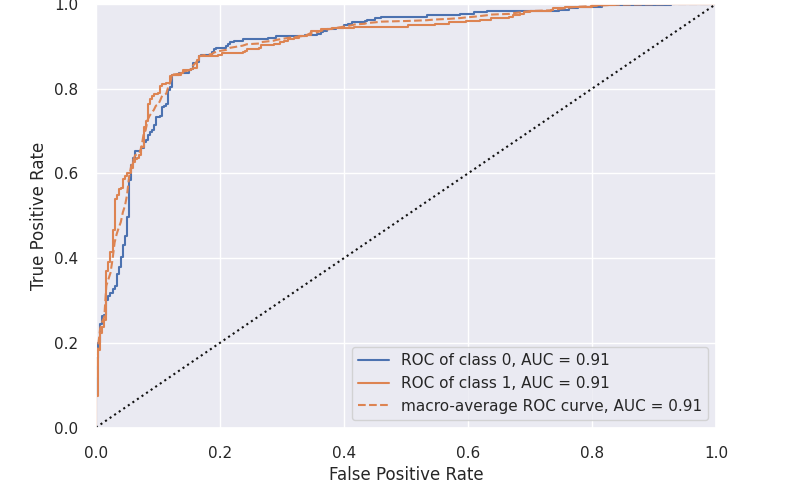}
\caption{Receiver Operating Curve of \textit{SuperCone}
on public benchmark data \texttt{madelon} for both classes
}
\label{fig-madelon auc}
\vspace{-10pt}
\end{figure}

We show the receiver operating curve for \texttt{a9a} dataset in \autoref{fig-a9a auc}
and the receiver operating curve for  \texttt{madelon} dataset in \autoref{fig-madelon auc},
further confirming the superior performance of \textit{SuperCone} of dataset with different difficulties.

\begin{figure}[]
\centering
\includegraphics[width=.45\textwidth]{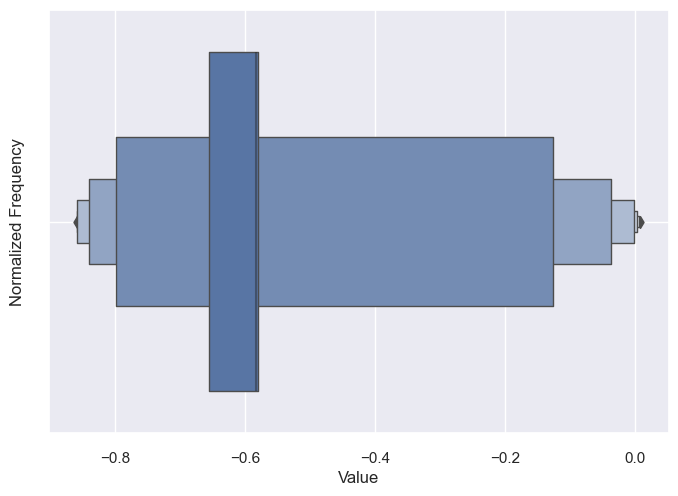}
\caption{Distribution of change in cross entropy log loss between \textit{SuperCone} and its ablated version 
}
\label{fig-log loss ablation}
\vspace{-10pt}
\end{figure}


\section{Details on the Ablation study}
\autoref{fig-log loss ablation} shows the difference between \textit{SuperCone} and its ablation version without heterogeneous experts in the commonly compared cross entropy log loss.
Specifically, the distribution of relative change in the log loss across all the predictive segments tasks are drawn, from which we can observe a consistent trend of loss reduction.

\section{Acknowledgement}
We acknowledge the helpful comments and engineering support from Akshay Gupta, Lisa Jones, 
Jason Grisby, Logan Palanisamy, Fei Tan in helping us develop the formulation and production solution.

    \begin{algorithm}[H]
    \caption{\textit{SuperAug} Algorithm}
    \label{alg:maml}
    \begin{algorithmic}[1]
    \REQUIRE 
label function of interest
$\mathbf{y}:\mathcal{S} \to \mathcal{Y}$, a 
 sampled dataset $\mathscr{D}\defeq \{\mathcal{D}^{train}, \mathcal{D}^{test}\}$
with each instance associated with concept vocabulary $\mathcal{C}$,heterogeneous experts $h_j$ with inner training oracle $\theta^*_j(\omega,\mathcal{D})$ for $j = 1\ldots J$
    \REQUIRE $K$: maximum depth for constructing experts, $V$: number of possible values for cross validation scheme
    \STATE  $\mathcal{D}^{train~(0)} \leftarrow \mathcal{D}^{train}$ 
    \FORALL{$k \in \{1\ldots K\}$}
        \FORALL{$s$ in $\mathcal{D}^{train}$}
        \STATE $V^{(k)}(s) \leftarrow$ random draw from $\{1\ldots V\}$
        \ENDFOR
        \STATE construct $\mathcal{D}^{train~(k)}$ according to \autoref{eq:cvtrain1}, \autoref{eq:cvtrain2} and \autoref{eq:cvtrain3}
    \ENDFOR
    \STATE obtain meta-trained $\omega^*$ according to \autoref{eq:metaloss}
        \FORALL{$k \in \{0\ldots K\}$}
            \FORALL{$j \in \{1\ldots J\}$}
                \STATE adapt experts $h^{(k)}_{j}$ from support  $\mathcal{D}^{train~(k,j)}_{target} \defeq  \mathcal{D}^{train~(k)}$ according to \autoref{eq:oracle}
            \ENDFOR
        \ENDFOR

    \STATE obtain final model based on the optimized meta parameter and adapted experts according to $\label{eq:final}$  
    \end{algorithmic}
    \end{algorithm}    
\end{document}